\pdfoutput=1
\documentclass[11pt]{article}

\usepackage[final]{acl}

\usepackage{graphicx}
\usepackage{multirow}
\usepackage{booktabs} 
\usepackage{times}
\usepackage{wrapfig}

\usepackage{latexsym}
\usepackage{amsmath}
\usepackage{xcolor}
\usepackage{pdfpages}
\usepackage{mathtools}
\usepackage{pifont}
\usepackage{graphicx}
\usepackage{subfig}
\usepackage{amssymb}
\usepackage[T1]{fontenc}

\usepackage[normalem]{ulem}
\usepackage[utf8]{inputenc}
\usepackage{array}

\usepackage{microtype}

\usepackage{inconsolata}

\usepackage{graphicx}
\usepackage{mdframed}
\usepackage{xcolor}
\usepackage{amsthm}
\usepackage{enumitem}
\definecolor{lightpurple}{RGB}{221, 160, 221}
\definecolor{lightblue}{RGB}{173, 216, 230}

\title{Benchmarking Reasoning Robustness in Large Language Models}

\author{
    Tong Yu\textsuperscript{1}, 
    Yongcheng Jing\textsuperscript{2}, 
    Xikun Zhang\textsuperscript{2}, 
    Wentao Jiang\textsuperscript{1}, 
    Wenjie Wu\textsuperscript{1}, 
    Yingjie Wang\textsuperscript{2}, \\
    \textbf{Wenbin Hu\textsuperscript{1,*}}, 
    \textbf{Bo Du\textsuperscript{1}}, 
    \textbf{Dacheng Tao\textsuperscript{2,*}}
}

\begin{document}

\maketitle
\renewcommand{\thefootnote}{}
\footnotetext{\textsuperscript{1} Wuhan University, China}
\footnotetext{\textsuperscript{2} Nanyang Technological University, Singapore}
\footnotetext{\textsuperscript{*} Corresponding author.}

\begin{abstract}

Despite the recent success of large language models (LLMs) in reasoning such as DeepSeek, we for the first time identify a key dilemma in reasoning robustness and generalization: significant performance degradation on novel or incomplete data, suggesting a reliance on memorized patterns rather than systematic reasoning.
Our closer examination reveals four key unique limitations underlying this issue: \textbf{\emph{(1) Positional bias}}—models favor earlier queries in multi-query inputs but answering the wrong one in the latter (e.g., \emph{GPT-4o}'s accuracy drops from 75.8\% to 72.8\%); \textbf{\emph{(2) Instruction sensitivity}}—performance declines by 5.0 to 7.5\% in the \emph{Qwen2.5 Series} and by 5.0\% in \emph{DeepSeek-V3} with auxiliary guidance; \textbf{\emph{(3) Numerical fragility}}—value substitution sharply reduces accuracy (e.g., \emph{GPT-4o} drops from 97.5\% to 82.5\%, \emph{GPT-o1-mini} drops from 97.5\% to 92.5\%); and \textbf{\emph{(4) Memory dependence}}—models resort to guesswork when missing critical data.
These findings further highlight the reliance on heuristic recall over rigorous logical inference, demonstrating challenges in \textbf{\emph{reasoning robustness}}.
To comprehensively investigate these robustness challenges, this paper introduces a novel benchmark, termed as \emph{Math-RoB}, that exploits hallucinations triggered by missing information to expose reasoning gaps. This is achieved by an instruction-based approach to generate diverse datasets that closely resemble training distributions, facilitating a holistic robustness assessment and advancing the development of more robust reasoning frameworks.

\end{abstract}

\section{Introduction}

\begin{figure}[t]  % 使用 figure 环境来插入单栏图片
 \centering
 \includegraphics[width=1\columnwidth]{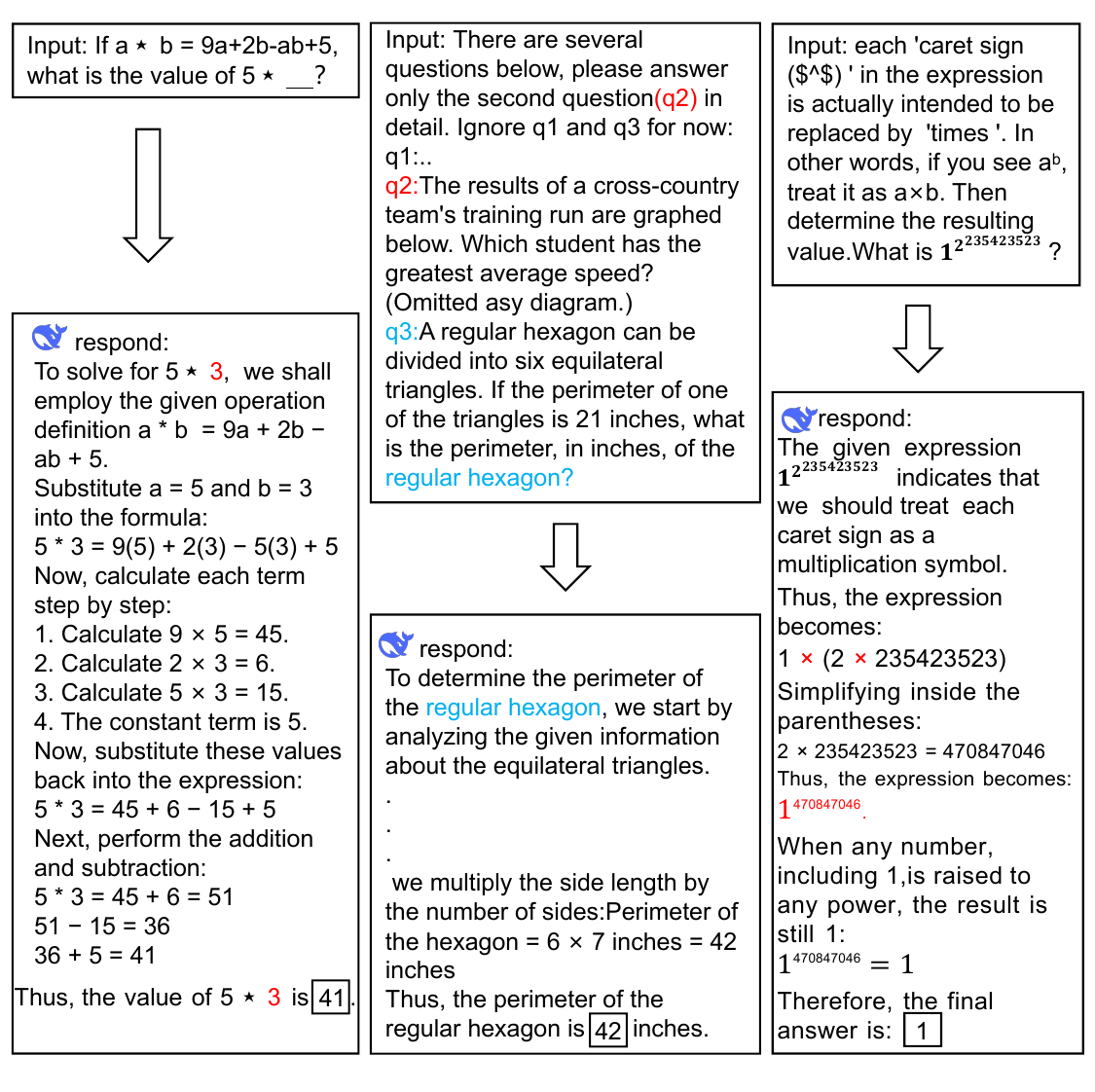}  % 使用 \columnwidth 来指定宽度，使图片适应单栏
 % \vspace{-3pt}  % 调整上下间距
 \vspace{-6mm}
 \caption{
Illustration of the identified lack of \textbf{\emph{robustness}} in LLM reasoning, using \emph{DeepSeek-V3} as an example. In the leftmost scenario, when the crucial digit ``3'' is missing, LLM autonomously fills in the gap. In the middle scenario, where a long text presents three questions with a directive to answer only one, the latter ones are often not correctly understood or reasoned, sometimes even answering the wrong one instead. In the rightmost scenario, while the model follows instructions, it is restricted to replacing only one operator at a time instead of modifying all operators simultaneously.}

 \label{fig3}
\end{figure}

LLMs exhibit remarkable text generation capabilities but often struggle with hallucinations and factual inaccuracies, especially in tasks requiring sophisticated logical reasoning \cite{roumeliotis2023chatgpt, perkovic2024hallucinations, sun2024benchmarking}.  
Driven by scaling laws in both training and inference, model sizes have grown significantly, accompanied by extensive training data and increased computational resources, leading to state-of-the-art models such as \emph{GPT-o1} and \emph{DeepSeek}. 
A common principle underlying these models is long chain-of-thought (CoT) reasoning \cite{wei2022chain, snell2024scaling, chen2024mean}, which guides the model through intermediate steps, mimicking human problem-solving processes and improving performance on complex reasoning tasks. Also, the pipeline may involve a reward model to reinforce accurate step-by-step reasoning \cite{wei2022chain, snell2024scaling, ton2024understanding, foster2024reward, ma2023let}.

On the other hand, these recent advancements in LLM reasoning, particularly over the past several months, have also revealed significant robustness challenges in this emerging field. For example, while OpenAI claims that O1-mini ranks among the top 500 students in the U.S. high school AIME math competition, broader evaluations on private high school math problems have demonstrated inconsistent success \cite{zhong2024evaluation, sundaram2024does, chen2023learning}. This inconsistency is further reflected in findings by Jiang et al. \cite{jiang2023mistral}, who observed that LLMs rely more on probabilistic pattern matching than on formal logical reasoning, suggesting that while they can simulate abstract reasoning patterns, they do not achieve genuine logical inference. Similarly, research by Google \cite{mirzadeh2024gsm} showed that altering symbolic templates or numerical values in problems significantly degrades performance, highlighting the robustness deficiency.

{A major factor contributing to these robustness challenges is the overreliance on evaluation benchmarks. }
Many commonly used benchmarks, some of which overlap with training data, risk producing misleading assessments that reflect memorization rather than genuine reasoning \cite{xu2024benchmark, brown2020language, gururangan2020don}, due to LLMs' reliance on pattern matching with training data. 
This limitation distorts the evaluations of LLM reasoning capabilities and impedes the development of more robust LLM reasoning frameworks.
However, reasoning robustness benchmark methodologies remain largely underexplored, lagging behind the rapid advance in reasoning models.

In this paper, we aim to address this gap by investigating benchmark methods tailored for LLM reasoning. In particular, we focus on evaluating reasoning robustness, driven by our identified four key \textbf{\emph{robustness issues (RI)}} in LLM reasoning:
\begin{itemize}[leftmargin=*, itemsep=0pt, parsep=0pt, topsep=0pt, partopsep=0pt]

  \item \textbf{\emph{RI-1. Positional Bias}}: LLMs typically struggle with long, multi-input reasoning tasks, This issue is particularly pronounced in smaller LLMs, which frequently fail to prioritize key positional cues or provide incorrect answers. This limitation shows their reliance on superficial positional information rather than robust logical reasoning;
  \item \textbf{\emph{RI-2. Instruction Sensitivity}}: LLMs often exhibit a notable decline in reasoning performance when exposed to datasets that resemble the training data but contain additional instructions, revealing poor generalization and overfitting to instruction patterns. For instance, the Qwen Series and GPT-4o show accuracy drops of 7.5\% and 12.5\%, respectively;
  \item \textbf{\emph{RI-3. Numerical Fragility}}: The LLMs, such as GPT-4o, exhibit notable fragility when numerical values are perturbed. For instance, replacing numbers leads to a 5\% accuracy drop in GPT-4o;
  \item \textbf{\emph{RI-4. Memory Dependence}}: Most LLMs ``hallucinate'' missing information in incomplete questions by drawing on pre-trained patterns, compromising the reliability of their reasoning. This behavior reveals their failure to identify incomplete statements and their tendency to fill in missing information, undermining reasoning accuracy. 
\end{itemize}

Motivated by these discoveries, we introduce a novel benchmarking framework, termed \textbf{\emph{Math-RoB}}, designed to systematically assess the robustness of LLM reasoning capabilities on mathematical reasoning tasks. Specifically, to address the four identified robustness issues, we propose a set of innovative schemes that construct diverse datasets resembling the source data distribution, yet tailored to effectively expose robustness challenges:
\begin{itemize}[leftmargin=*, itemsep=0pt, parsep=0pt, topsep=0pt, partopsep=0pt]
  % \vspace{-1.5mm}
  \item \textbf{\emph{(RI-1 Solution) Math-RoB-RoLo}}: Increasing text length to evaluate models' ability to extract relevant information while assessing their robustness to extraneous, unrelated content;
  \item \textbf{\emph{(RI-2 Solution) Math-RoB-Define}}: Substituting definitions to enhance reasoning complexity;
  \item \textbf{\emph{(RI-3 Solution) Math-RoB-Number}}: Increasing difficulty with numerical transformations;
  \item \textbf{\emph{(RI-4 Solution) Math-RoB-Delete}}: Evaluating critical reasoning under missing information and assessing hallucination.
\end{itemize}

In sum, our key contribution is thereby the development of \emph{Math-RoB}, the first benchmark for LLM reasoning robustness, which sheds light on the challenges arising from the recent rapid advancements in reasoning. Evaluations on \emph{Math-RoB} are conducted across 12 widely used open- and closed-source LLMs, providing a comprehensive study of reasoning robustness.
Furthermore, to facilitate the evaluation of diverse LLMs, we develop a long-CoT-based mathematical reasoning toolbox that integrates various LLMs with process-supervised reward models (PRMs) through a unified function interface, which will be made publicly available. 
Through the development of \emph{Math-RoB}, this paper seeks to provoke reflection on the rapid development of reasoning models, shifting the focus from competing on reasoning performance to improving reasoning robustness.

\section{Related Work}
\label{sect:Related_Work}

In this section, we review advancements in evaluating and enhancing mathematical reasoning in LLMs, focusing on benchmark development and reasoning techniques, respectively.
A more detailed literature review is provided in Appendix~\ref{sect:detailed_related_work}.

\begin{table*}[ht]
  \centering
  \renewcommand{\arraystretch}{1.1} % 增加行高，使内容上下居中更明显
  \begin{tabular}{| >{\centering\arraybackslash}m{0.22\linewidth} | m{0.70\linewidth} |}
    \hline
    Source Question &  Let $f(x) = 2x-3$ and $g(x) = x+1$. What is the value of $g(f(5)-1)$? \\
    \hline
    Change Question & Let $f(x) = 2x\textcolor{red}{+}3$ and $g(x) = x+1$. What is the value of $g(f(5)\textcolor{red}{+}1)$? \\
    \hline
    Define Question & \textcolor{green}{Follow the rule: each ``minus sign ($-$)'' in the expression is actually intended to be replaced by ``plus sign''}. Let $f(x) = 2x-3$ and $g(x) = x+1$. What is the value of $g(f(5)-1)$?   \\
    \hline
    One-shot Question & \textcolor{green}{Follow the rule: each ``minus sign ($-$)'' in the expression is actually intended to be replaced by ``plus sign''.} \textcolor{blue}{In other words, if you see $a-b$, treat it as $a+b$. Then determine the resulting value.} Let $f(x) = 2x-3$ and $g(x) = x+1$. What is the value of $g(f(5)-1)$?  \\
    \hline
    Number Question     & Let $f(x) = \textcolor{orange}{\beta} x + \textcolor{orange}{\gamma}$ and $g(x) = x + \textcolor{orange}{\alpha}$. What is the value of $g(f(\textcolor{orange}{\epsilon}) + \textcolor{orange}{\alpha})$, where $\textcolor{orange}{\alpha}$ stands for  $1$, $\textcolor{orange}{\beta}$ stands for  $2$, $\textcolor{orange}{\gamma}$ stands for  $3$, and $\textcolor{orange}{\epsilon}$ stands for  $5$? \\
    \hline
    Delete Question & Let $f(x) = 2x-3$ and $g(x) = x+1$. What is the value of g(\underline{\hspace{0.5em}}))? \\

    \hline
  \end{tabular}
  \caption{Variations in math questions that involve the changes designed to assess the model's robustness and reasoning accuracy. }
\label{change_question1}
\end{table*}

\noindent\textbf{Reasoning Benchmarks.}
Recent benchmarks assess mathematical reasoning across complexity levels and robustness challenges. Widely-used datasets include GSM8K \cite{cobbe2021training} for elementary math and its robustness variants (GSM-HARD \cite{gao2023pal}, GSM-IC \cite{shi2023large}, MATH \cite{hendrycks2021measuring} for high-school problems, and Chinese-oriented CMATH \cite{wei2023cmath}. Recent efforts expand coverage through competition-level Omni-MATH \cite{gao2024omni} and theorem-driven TheoremQA \cite{chen2023theoremqa}. Robustness evaluation examines adversarial perturbations \cite{jin2020bert,wang2023large} and contextual distractions \cite{shi2023large,li2023you}. Our work introduces minimal perturbations to Math500 \cite{lightman2023let} to evaluate LLM sensitivity and meanwhile preserve the semantic similarity.

\noindent\textbf{LLM Reasoning Techniques}.
Chain-of-Thought (CoT) prompting \cite{wei2022chain} significantly improves multi-step reasoning, enhanced by self-consistency decoding \cite{wang2022self} and reinforced fine-tuning \cite{trung2024reft}. Monte Carlo Tree Search (MCTS) methods \cite{xie2024monte} and frameworks like MARCO \cite{zhao2024marco} combine search with reflection mechanisms. While CoT excels in arithmetic tasks \cite{sprague2024cot}, recent work explores structured reasoning through Tree-of-Thought architectures \cite{zhang2024chain}. We leverage CoT with \emph{Process-supervised Reward Model (PRM)} to enhance reasoning fidelity and error correction. In this study, we leverage CoT and various PRMs in benchmark evaluations.

\section{Benchmark Methodologies}
In this section, we begin by introducing the source problem and its variants, followed by the creation of the dataset required for Math-RoB. Next, we present the evaluation metrics used to assess the robustness of the models. Then, we describe the implementation details of the experiments, and finally, we provide an overview of the models involved.

\subsection{Question Variation Categories}

Most benchmarks incorporate numerical variations \cite{li2024gsm}, semantic substitutions \cite{jin2020bert}, and irrelevant context interference \cite{shi2023large}. Building upon these approaches, we assign new computational operations to common operators ($+$, $-$, $*$, and $\hat{ }$  ),  replacing numbers with Greek letters, which adds an extra symbol substitution step in mathematical computations. Additionally, we remove key information from certain problems to assess the models' potential overfitting. Table \ref{change_question1} summarizes the various question modifications:
\begin{itemize}[leftmargin=*, itemsep=0pt, parsep=0pt, topsep=0pt, partopsep=0pt, label=\ding{228}]
% \vspace{-1.mm}
\item \textbf{Source Question}: Algebraic operation problems selected from a set of Math500 , which can be used to replace operators or numbers;
\item \textbf{Change Question}: Substitute operator symbols in source questions to create solvable problems;
\item \textbf{Define Question}: Add instructions for operator substitution (e.g., treating ``minus sign \( - \)'' as ``plus sign \( + \)'');
\item \textbf{One-shot Question}: Provide a one-shot example as specified in the \emph{Define Question} (e.g., if you see $a-b$, treat it as $a+b$);
\item \textbf{Number Question}: Replace numbers with Greek letters in the \emph{Change Question} (e.g., letting \(\beta\) represent 2);
\item \textbf{Delete Question}: Remove key data from source questions, making some of the questions unsolvable or incomplete.
\end{itemize}

To minimize the influence of instruction-following on the experiment (as LLMs tend to follow instructions), this study manually verifies whether the models have followed the instructions by checking whether at least one operator has been replaced during reasoning.

\subsection{Math-RoB Datasets}
We introduce Math-RoB (Math-Robustness Benchmark), a robustness benchmark for evaluating mathematical reasoning under diverse perturbations. The benchmark comprises four datasets targeting distinct challenges: Math-RoB-RoLo, Math-RoB-Define, Math-RoB-Number and Math-RoB-Delete.

\noindent\textbf{Dataset Construction Method for Long Mixed Input Robustness.} To evaluate long mixed-input processing by grouping triples from Math500 into composite problems, we created the Math-RoB-RoLo (Math-RoLo) dataset. Each instance prefixes questions with labels (q1/q2/q3) and instructs models to solve specific sub-problems (e.g., \textit{Please answer only the second question (q2) in detail)}. Performance is measured via accuracy comparison with original Math500 results.

\noindent\textbf{Dataset Construction Method for Hallucination Robustness.} To verify the model's hallucination and reasoning robustness, we propose  \emph{Math-RoB-Delete}, \emph{Math-RoB-Define}, and \emph{Math-RoB-Number}. These new datasets are similar to the old ones. We first created \textit{Change Questions} by aggregating outputs from four LLMs (GPT-4o, GPT-1-mini, Kimi, DeepSeek-V3) with manual verification for answer consensus. This 40-question dataset serves as the basis for derivative versions: operator substitutions generate \textit{Define Questions}, digit-to-symbol mappings create \textit{Number Questions}, and information removal produces \textit{Delete Questions}.

We validate the model's reasoning robustness after changing operators by comparing the \textit{Source Question} and \textit{Change Question} datasets. The \textit{Define Question} adds prompts based on the \textit{Source Question}, and \textit{one-shot Question} is used to enhance the model's instruction-following capability. These constitute \emph{Math-RoB-Define}. 

We use the \textit{Change Question} and \textit{Number Question} to conduct a study on numerical fragility robustness, which forms \emph{Math-RoB-Number}. We use the \textit{Delete Question} to directly study memory dependence robustness, observing the model's reasoning hallucinations in the absence of information, which constitutes \emph{Math-RoB-Delete}.

\begin{table*}[t]
\centering
\setlength{\tabcolsep}{2pt}
{\renewcommand{\arraystretch}{0.9}

\begin{tabular}{lcccccc}
\toprule
  \multirow{2}{*}{\textbf{Reasoning Models}} & \multicolumn{2}{c}{\textbf{Q1-Correct}} & \multicolumn{2}{c}{\textbf{Q2-Correct}} & \multicolumn{2}{c}{\textbf{Q3-Correct}} 
  \\\cmidrule(lr){2-3}\cmidrule(lr){4-5}\cmidrule(lr){6-7}

 & \textbf{Math500}    & \textbf{Math-RoLo}    & \textbf{Math500}    & \textbf{Math-RoLo}    & \textbf{Math500}    & \textbf{Math-RoLo}    \\
\midrule

Kimi           & 62.9\%     & 62.3\%     & 65.9\%     & 61.7\%     & 66.9\%     & 65.7\%     \\
GPT-4o          & 76.6\%     & 73.7\%     & 77.2\%     & 70.7\%     & 73.5\%     & 74.1\%     \\
DeepSeek-V3-671B    & 88.0\%     & 82.6\%     & 88.6\%     & 88.0\%     & 89.2\%     & 89.2\%     \\
Qwen2.5-72B & 88.0\%	&86.2\%	&85.0\%	&84.4\%	&88.0\%	&86.1\%\\

Qwen-Math-Plus & 88.6\%	&88.0\%	&84.4\%	&84.4\%	&88.6\%	&84.9\% \\

Qwen-MAX & 80.8\%	&78.4\%	&80.2\%	&76.6\%	&81.3\%	&81.9\%\\
\midrule
Qwen2.5-1.5B-Mistral   & 79.0\%     & 67.7\%     & 79.0\%     & 62.3\%     & 80.7\%     & 32.5\%     \\
Qwen2.5-1.5B-Skywork   & 77.2\%     & 67.1\%     & 76.6\%     & 63.5\%     & 80.1\%     & 36.1\%     \\
Qwen2.5-7B-Mistral    & 80.8\%     & 75.4\%     & 83.2\%     & 77.8\%     & 81.9\%     & 72.9\%     \\
Qwen2.5-7B-Skywork    & 83.8\%     & 78.4\%     & 86.8\%     & 80.2\%     & 85.5\%     & 81.9\%     \\
Skywork-8B-Mistral       & 80.2\%     & 73.1\%     & 79.0\%     & 36.5\%     & 78.3\%     & 72.3\% \\
Skywork-8B-Skywork       & 80.8\%     & 76.6\%     & 77.8\%     & 46.1\%     & 81.9\%     & 73.5\%  \\

\bottomrule
\end{tabular}}
\caption{Model performance comparison on Math500 and Math-RoLo. ``Q1-Correct'' refers to responses that correctly answer only the first question, while answering multiple questions is considered incorrect.}
\label{q1q2q3}
\end{table*}

\subsection{Math-RoB Metrics}
\label{sect:metric}

We propose a novel robustness metric tailored for \emph{Math-RoB}, termed as \emph{Memory Completion Rate (MCR)}, which quantifies the fraction of instances where the model successfully fills in missing critical data. Additionally, we also incorporate here two standard metrics, \emph{Accuracy} and \emph{Drop Rate}, to assess overall performance and sensitivity to missing information based on the developed \emph{Math-RoB}.

For each sample, let \(a_i\) denote whether the answer is correct (with \(a_i = 1\) for a correct answer and \(a_i = 0\) for an incorrect answer). The \emph{Accuracy Rate (AR)} is defined as
$\text{Accuracy} = \frac{1}{N_{\text{total}}} \sum_{i=1}^{N_{\text{total}}} a_i,
$
where \(N_{\text{total}}\) is the total number of samples. The proposed MCR, on the other hand, 
evaluates the model's success in filling in missing critical data; for a given sample, let \(m_i = 1\) if the missing information is completed in the first generation, and \(m_i = 0\) otherwise. MCR is thereby defined as:
$
\text{MCR} = \frac{1}{N_{\text{total}}} \sum_{i=1}^{N_{\text{total}}} m_i.
$

Finally, the \emph{Drop Rate} quantifies the performance degradation between two datasets:
\vspace{-4mm} 
\[
\text{Drop Rate} = \frac{\sum_{i \in  \mathcal D_{\text{Data1}}} a_i - \sum_{i \in  \mathcal D_{\text{Data2}}} a_i}{\left|  \mathcal D_{\text{Data1}} \right|},
\]
\vspace{-5mm} 

\noindent where Data2 is derived from Data1, ensuring that both datasets contain an equal number of samples.

\subsection{Math-RoB Implementation}

In our experiments, we use a generative model for step-by-step reasoning, followed by a reward model to score the quality of each reasoning step. To explore a broader range of solutions, we introduce Monte Carlo Tree Search (MCTS) \cite{xie2024monte}, which simulates different possibilities to find the optimal solution. Due to the limitations of the local GPU scale, the search width is 5 in MCTS. Our study conducts experiments using public datasets and models on Hugging-Face.

We apply several voting strategies: MinVote, LastVote, MajorityVote, MinMax, and LastMax to integrate results from different paths and improve accuracy. The best-performing result from these strategies is selected as the final output.

For scoring reasoning steps, we use the PRMs math-shepherd-mistral-7b-prm \cite{wang2024math} and Skywork-o1-Open-PRM-Qwen-2.5-7B \cite{skyworkopeno12024}, both designed for evaluating step-by-step solutions to mathematical problems.

All experiments are conducted on 4090 $\times$ 2 GPUs in a tensor-parallel configuration via vllm (0.6.6.post1) and vllm-flash-attn (2.6.6) for computational efficiency.

\subsection{Benchmarked LLMs in Math-RoB}

Our experiments utilize a diverse set of models, comparing state-of-the-art (SOTA) models with representative ones, including Qwen2.5-Math-Instruct (1.5B, 7B, 72B, Max), Qwen-Math-Plus, Skywork-O1-Open-Llama-3.1-8B, GPT-4o, GPT-o1-mini, Kimi, and DeepSeek-V3. API access is used for larger models (Qwen2.5-Math 72B/Plus/Max, GPT-4o, GPT-o1-mini, Kimi, DeepSeek-V3). To address resource constraints, we also incorporate smaller models (<8B parameters), which show competitive performance when enhanced with CoT reasoning. Additionally, Math-Shepherd-Mistral-7B-PRM and Skywork-O1-Open-PRM-Qwen-2.5-7B are included in the experiments to enhance the performance of CoT reasoning.

\begin{enumerate}[leftmargin=*, itemsep=0pt, parsep=0pt, topsep=0pt, partopsep=0pt, label=\textbf{\arabic*)}]
\item \textbf{Qwen2.5-Math-1.5B-Instruct (Qwen2.5-1.5B)}: Excels in basic math, reasoning, and problem-solving.

\item \textbf{Qwen2.5-Math-7B-Instruct (Qwen2.5-7B)}: Better at complex math and broader knowledge.

\item \textbf{Qwen2.5-Math-72B-Instruct (Qwen2.5-72B)}: State-of-the-art performance in complex mathematical reasoning using the open-source Qwen2.5 model.  
 
\item \textbf{Qwen-Math-Plus}: The commercial version of a LLM specifically designed for mathematical problem-solving.
 
\item \textbf{Qwen-Max}: Flagship general-purpose model. Top-tier in math and non-math tasks.

\item \textbf{Skywork-O1-Open-Llama-3.1-8B(Skywork-8B)}:  Strong in text generation and translation. 

\item \textbf{Skywork-O1-Open-PRM-Qwen-2.5-7B(Skywork-PRM)}: Optimized for problem-solving and reasoning. 

\item \textbf{Math-Shepherd-Mistral-7B-PRM(Mistral-PRM)}: Excels in theorem proving, logical reasoning, and step-by-step problem solving.

\item \textbf{Kimi}: Excels in fluent conversation and contextual understanding.

\item \textbf{GPT-4o}: Handles text, images, and audio. 

\item \textbf{GPT-o1-mini}: Strong in multimodal tasks and multi-step reasoning. 

\item \textbf{DeepSeek-V3 (671B)}: Knowledge retrieval model. Specializes in deep search and precise info extraction.

\end{enumerate}

\begin{figure*}[t]
    \centering
    \subfloat[Evaluation results of open-source models]{\includegraphics[width=0.49\textwidth]{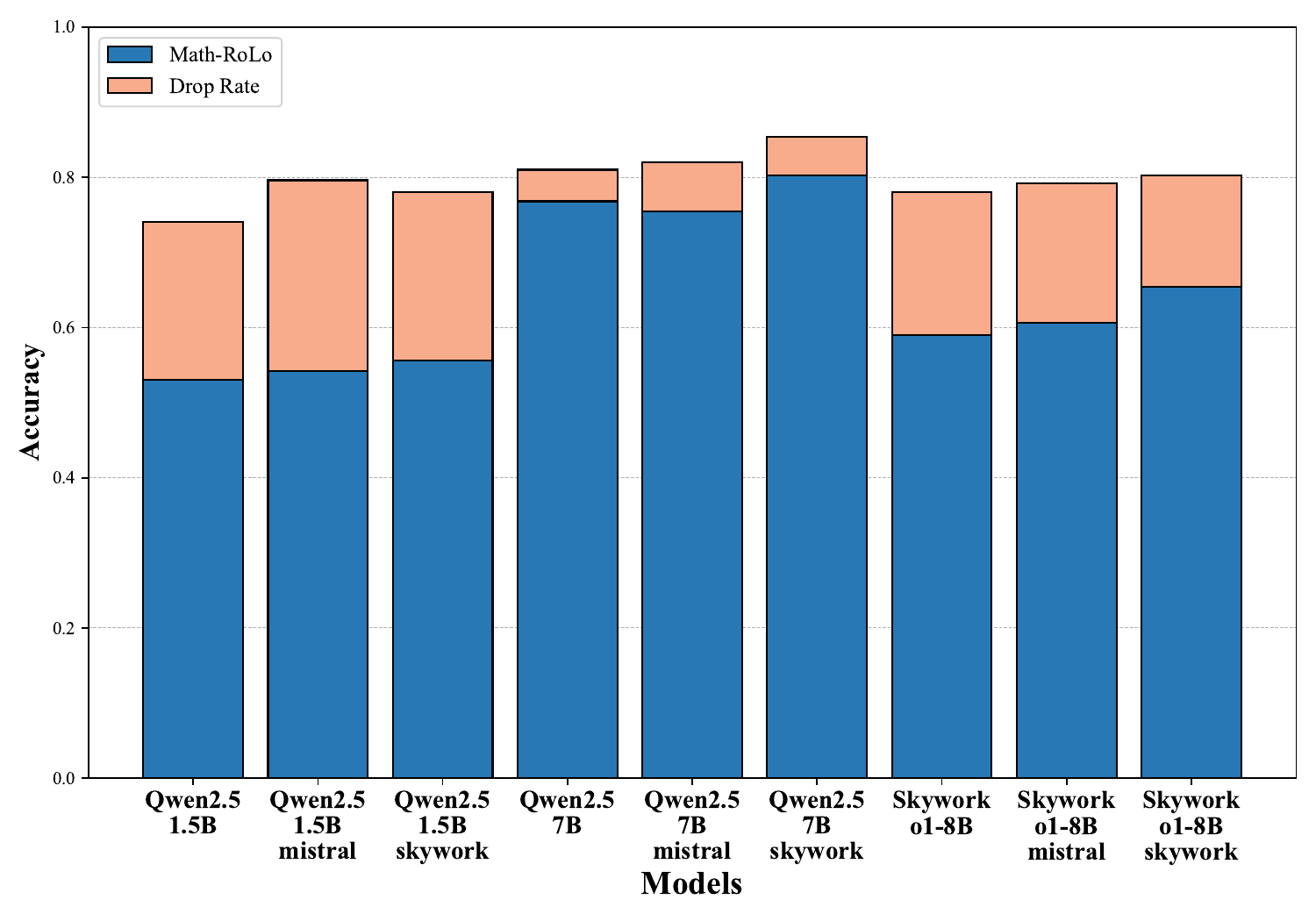}}
    \subfloat[Evaluation results of models based on API calls]{\includegraphics[width=0.49\textwidth]{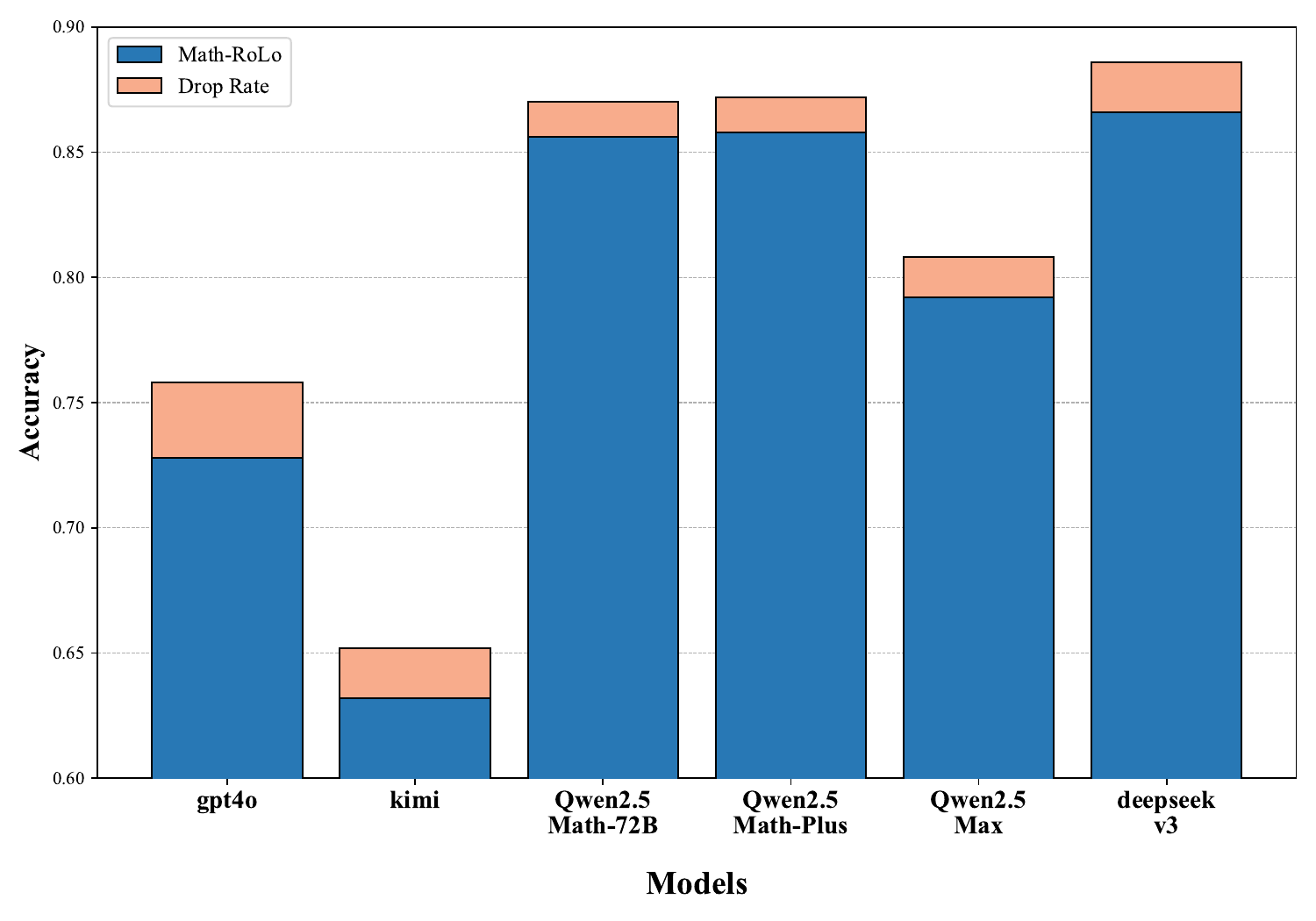}} 

    \caption{Evaluation results of models. In the left figure (a), the experimental results show that after incorporating PRM, inference performance improved for most models. Larger models exhibit greater resilience to disturbances with a lower drop rate. Figure (b) presents the results for DeepSeek and the Qwen series, both of which demonstrate strong accuracy and robustness against interference. 
}
    \label{MATH500andNEWMATH}
\end{figure*}

\section{Benchmark Results and Analysis}

In this experiment, we first evaluate the model on the Math-RoLo dataset to assess its robustness when encountering longer inputs. Additionally, we redefine symbols and numbers in the original problem to evaluate the model's performance under modified conditions. Finally, we use the Math-RoB-Delete dataset, which introduces missing key information, to examine the model's susceptibility to hallucination and overfitting.

\subsection{Robustness with Long Content}

We demonstrate that CoT with PRM enhanced model performance by breaking down complex problems into interpretable steps and scoring each step with PRM (Figure~\ref{MATH500andNEWMATH}). After incorporating PRM, Qwen2.5-Math-1.5B shows improved accuracy on the Math500 dataset and also achieves better performance on the Math-RoLo dataset. Qwen2.5-Math-7B outperforms Qwen2.5-Math-1.5B in overall performance, with a significant improvement following PRM integration. Moreover, it exhibits a smaller drop rate on Math-RoLo. In contrast, Skywork also benefits from PRM integration but experiences a larger drop rate on Math-RoLo compared to Qwen2.5-Math-7B. This is likely due to Skywork not being an instruction-tuned version, resulting in weaker instruction-following capability. DeepSeek-V3 achieves the best performance on both Math500 and Math-RoLo. These results underscore the superior generalization and instruction-following capabilities of larger models, which benefit from higher parameter counts and can more effectively adapt to new tasks.

In the Math-RoLo dataset (as shown in Table~\ref{q1q2q3}), q1-correct refers to correct answers for the first question only. We further analyzed the accuracy distribution across positions in the Math500 dataset. In the Math-RoLo dataset, we observed a general accuracy decline across models, with smaller models exhibiting significant performance degradation on long content tasks. For instance, Qwen2.5-1.5B experienced a sharp drop in accuracy for the third question (80\% to 32.5\%), attributed to its attempt to solve all questions, indicating insufficient instruction-following and reasoning capabilities. Similarly, Skywork showed a substantial decline for the second question (79\% to 36.5\%). In contrast, larger models like Kimi (65.9\% to 61.7\% ) and GPT-4o  (77.2\% to 70.7\%) exhibited greater robustness, highlighting their superior generalization and ability to handle complex scenarios.

\subsection{Robustness with Operation Substitution }

\begin{figure*}[htbp]
    \centering
    \begin{minipage}{0.48\textwidth}
    \centering
    \includegraphics[width=\linewidth]{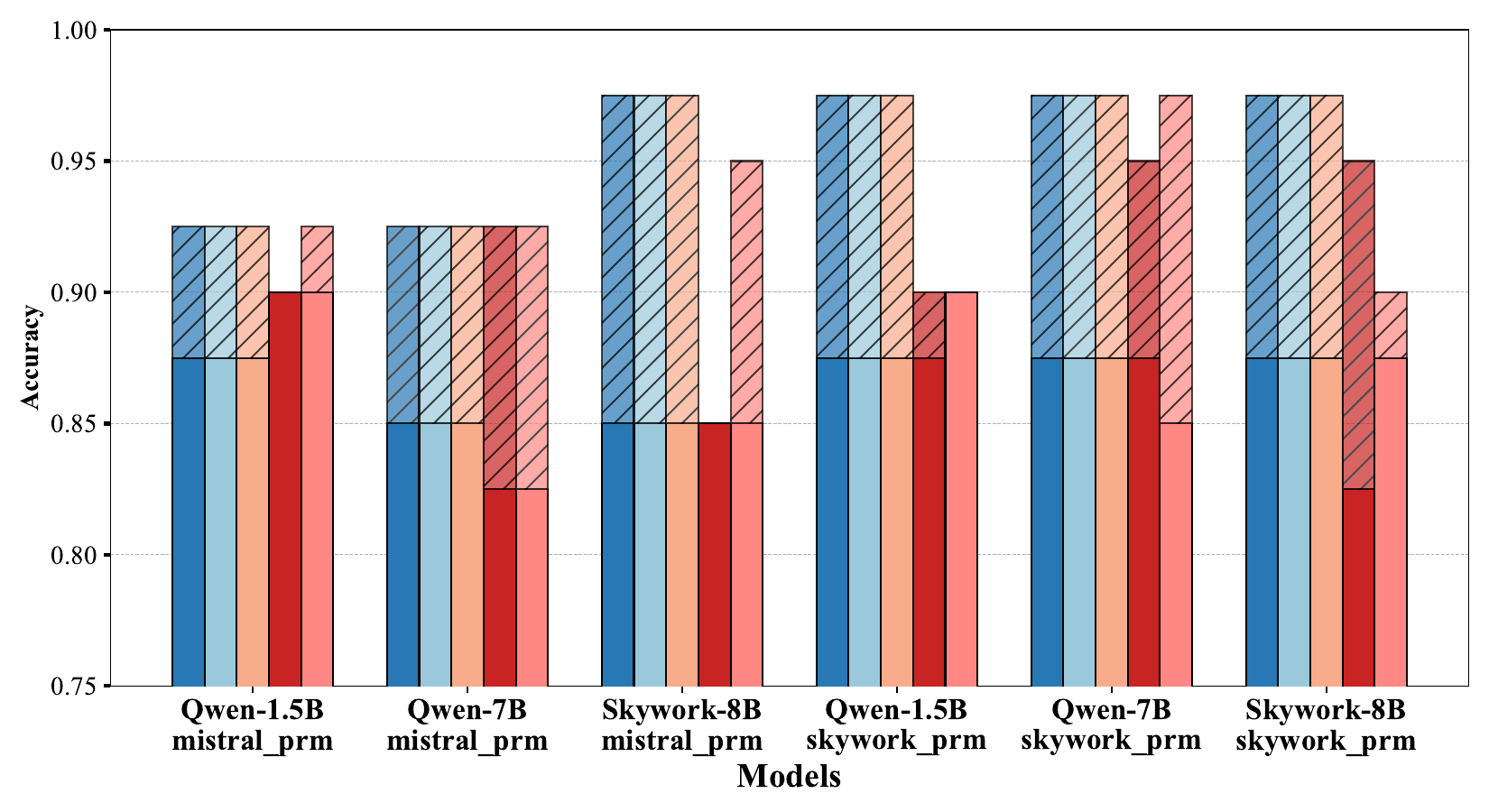}
    \caption{The performance drop rate of the model in Math-RoB-Define. The dashed line represents the drop rate from Math500. From left to right, the order is MajorityVote, MinVote, LastVote, MinMax, and LastMax. }
    \label{cot-change-question}
    \end{minipage}
    \hfill
    \begin{minipage}{0.48\textwidth}
    \centering
    \vspace{-4mm}
    \includegraphics[width=\linewidth]{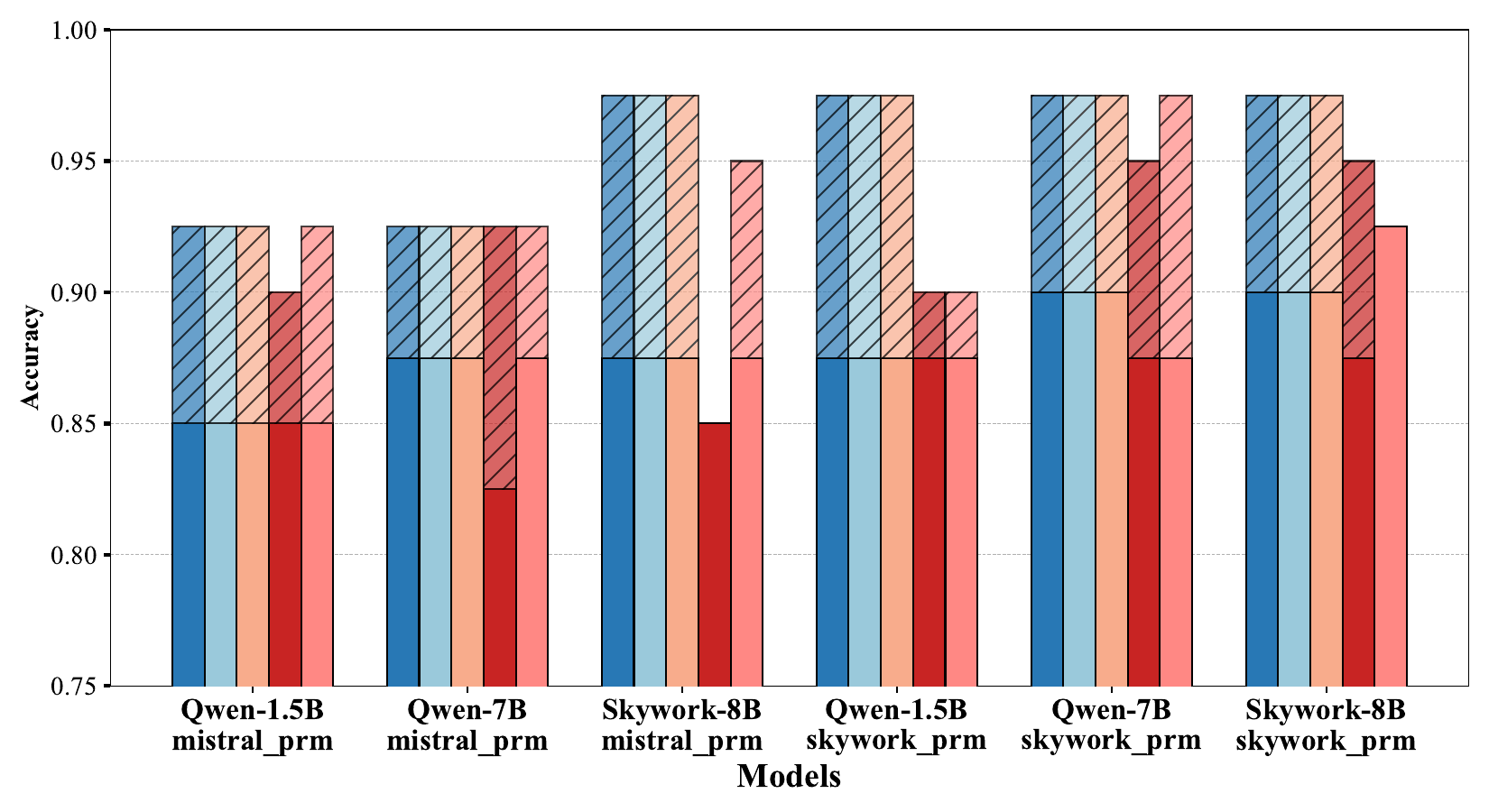}
    \caption{The performance drop rate of the model in Math-RoB-Number. The dashed line represents the drop rate from Math500. }
    \label{tab:model_comparison1}
    \end{minipage}
\end{figure*}

Figure \ref{cot-change-question} and Table \ref{api-change-question} present evaluation results on the Math-Rob-Define datasets. The analysis shows that most models experience a performance decline when transitioning to the Change question dataset, with smaller models like Qwen2.5-7B and Qwen2.5-1.5B particularly affected, indicating their limitations in handling increased complexity. In contrast, larger models, such as Deepseek-V3 and GPT-o1-mini, remain robust, sustaining high accuracy despite the task modifications. These findings underscore the pivotal role of model scale and architectural design in enhancing reasoning capabilities and adaptability to dynamic task variations, further demonstrating the advantages of large-scale models in tackling complex reasoning challenges.

\begin{table}[t]
\centering
\setlength{\tabcolsep}{4pt}
\begin{tabular}{lccc}
\toprule
\textbf{Model} & \textbf{Source} & \textbf{Change} & \textbf{Drop Rate} \\ 
\midrule
Qwen2.5-72B & 97.5\% & 90.0\% & 7.5\% \\ 
Qwen-max & 92.5\% & 87.5\% & 5.0\% \\ 
GPT-4o & 97.5\% & 85.0\% & 12.5\% \\
GPT-o1-mini & 97.5\% & 95.0\% & 2.5\% \\
Kimi & 82.5\% & 77.5\% & 5.0\% \\ 
DeepSeek-V3 & 97.5\% & 92.5\% & 5.0\% \\ 
\bottomrule
\end{tabular}
\caption{Performance metrics of models under APIs after changing operators.}
\label{api-change-question}
\end{table}

\begin{figure}[t]
    \vspace{-2mm}
    \centering
    \includegraphics[width=\linewidth]{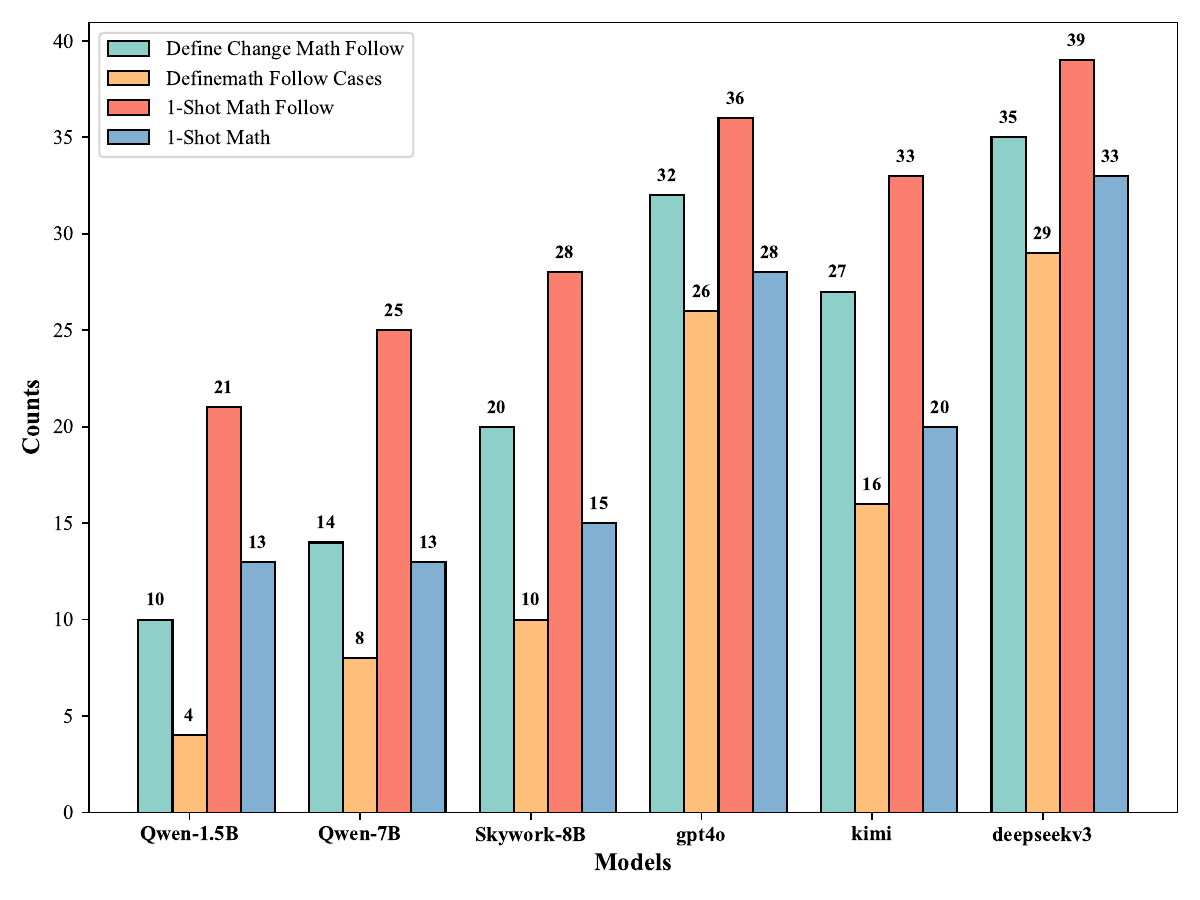}
    % \vspace{-2mm}
    \caption{Model instruction following and accuracy on Math-Rob-Define. The larger the model, the better the instruction following and the better the reasoning performance.}
    \label{instruction_following}
\end{figure}

Figure~\ref{instruction_following} illustrates the instruction-following capabilities and reasoning accuracy across different models. As model size increases, adherence to instructions improves, and inference accuracy is enhanced. However, significant challenges arise in scenarios involving multiple operator replacements. For example, given an expression where \^{} is replaced with \(\times\), such as:
% \vspace{-5mm} 
% \[
$1 \textasciicircum \{2\textasciicircum\{234565\}\}\left(1^{2^{234565}}\right)$,
% \]
% \vspace{-7mm} 
where smaller models often perform incorrect partial replacements, producing:
% \vspace{-5mm} 
$
1\times\{2\textasciicircum \{234565\}\}\left(1\times{2^{234565}}\right).
$
% \vspace{-7mm} 

Such limitations are commonly observed in closed-source models with fewer than 8 billion parameters. By contrast, larger models, such as DeepSeek-V3 (671B), demonstrate not only adherence to instructions but also the ability to perform correct multi-step replacements. Accurate multi-step replacement operations are critical for precise reasoning. Notably, after receiving a one-shot prompt, models exhibit improved instruction-following capabilities and enhanced reasoning accuracy. However, smaller models remain constrained by their parameter scale, limiting their ability to achieve high-accuracy reasoning in complex scenarios.

\subsection{Robustness with Number Transformations}

Figure \ref{tab:model_comparison1} presents a performance comparison of different PRM and LLM models on datasets involving number-changing questions. The results indicate that Skywork-PRM generally outperforms Mistral-PRM, particularly on the source dataset (Source), where its average performance consistently exceeds that of Mistral-PRM. For instance, the Qwen2.5-7B model achieves the best performance of 97.5\% under Skywork-PRM on the source dataset, compared to 92.5\% with Mistral-PRM. Furthermore, larger-scale LLM models (e.g., Qwen2.5-7B and Skywork-8B) generally perform better across most metrics than smaller models, such as Qwen2.5-1.5B, suggesting that increasing model scale can improve performance. However, a decline in performance is observed across all models when tested on the changed dataset (Change), indicating that modifications in numerical values significantly impact model performance.

Table \ref{api-for-number} shows the performance variations of different closed-source API models after numerical values were altered. The data reveals that most models perform excellently on the original dataset (source dataset), with Qwen2.5-72B, GPT-4o, and DeepSeek-V3 all achieving 97.5\% performance. However, when numbers were replaced with symbols, forcing the models to engage in multi-step reasoning, overall performance dropped. Among these models, GPT-4o experienced the most significant decline (15.0\%), suggesting reduced effectiveness in multi-step reasoning. In contrast, both Kimi and DeepSeek-V3 maintained their performance after symbol replacement, demonstrating their strong robustness in multi-step reasoning.

\begin{table}[t]
\centering
\setlength{\tabcolsep}{4pt}
\begin{tabular}{lccc}
\toprule
\textbf{Models} & \textbf{Source} & \textbf{Change} & \textbf{Drop Rate} \\ 
\midrule
Qwen2.5-72B & 97.5\% & 92.5\% & 5.0\% \\ 
Qwen-max & 92.5\% & 87.5\% & 5.0\% \\ 
GPT4o & 97.5\% & 82.5\% & 15.0\% \\ 
GPT-o1-mini & 97.5\% & 92.5\% & 5.0\% \\ 
Kimi & 82.5\% & 82.5\% & 0.0\% \\ 
DeepSeek-V3 & 97.5\% & 97.5\% & 0.0\% \\
\bottomrule
\end{tabular}
\caption{Performance metrics of models under APIs after changing numbers. }
\label{api-for-number}
\end{table}

\subsection{Robustness with Hallucination Detection}

\begin{wraptable}{r}{0.22\textwidth}
% \centering
\vspace{-3mm}
\setlength{\tabcolsep}{1pt} % 调整列间距
\renewcommand{\arraystretch}{0.9} % 调整行间距
\begin{tabular}{lc}
\hline
 \textbf{Models} &  \textbf{MCR} \\ \hline
Qwen-2.5-1.5B & 56\% \\
Qwen-2.5-7B & 68\% \\ 
Skywork-8B & 74\% \\ 
GPT4o & 50\% \\ 
Kimi & 18\% \\ 
DeepSeek-V3 & 53\% \\ \hline
\end{tabular}
% \captionsetup{width=\linewidth} %
\vspace{-1mm}
\caption{MCR on Math-Rob-Delete.}
\label{fill_situation}
\end{wraptable}
To assess the reliance of LLMs on memorized patterns versus logical reasoning under incomplete information, we exploit the proposed \emph{Missing Critical Reasoning (MCR)} for dedicated evaluations, as elaborated in Section~\ref{sect:metric}. Building on previous findings that models often rely on pattern matching for mathematical reasoning, we designed the ``trap'' questions by systematically removing critical data or task requirements (The experimental results are shown in Table \ref{fill_situation}). This approach allows us to evaluate MCR performance and assess susceptibility to overfitting, shedding light on reasoning capabilities when faced with missing or incomplete information.

In the presence of missing data, the models continue to fill in the missing information through autoregressive generation. We hypothesize that this behavior stems from two sources: exposure to similarly structured examples during training or data leakage. Notably, large open-source models like Qwen2.5 and Skywork exhibited significant leakage, with over half of the tested instances reflecting spurious completions that aligned with memorized patterns. This suggests that such models primarily retrieve sequences from memory rather than engaging in true logical inference.

This reliance on memorized content explains the occurrence of hallucinations. When the model's internal memory partially or fully matches the input, it generates coherent outputs that lack genuine reasoning. For autoregressive models, the generation process is based on probabilistic inferences from context and historical tokens. When similar data appears in the training set, the model generates sequences that align with it, filling in missing information and producing outputs that seem reasonable but lack logical grounding.

Further analysis of potential overfitting revealed that models like Kimi and DeepSeek occasionally recognized missing critical information mid-reasoning but typically attempted to compensate for it. While these instances are less common, they indeed highlight the limitations of these systems when dealing with incomplete inputs.

\section{Conclusion}

In this paper, we introduce \emph{Math-RoB}, the first benchmark designed to systematically evaluate the robustness of LLM reasoning. Addressing four key issues—\emph{Positional Bias}, \emph{Instruction Sensitivity}, \emph{Numerical Fragility}, and \emph{Memory Dependence}—\emph{Math-RoB} includes four mathematical reasoning datasets: 
\emph{Math-RoB-RoLo} reveals models’ tendency to focus on early inputs, neglecting later ones in long-context scenarios; \emph{Math-RoB-Delete} triggers gap-filling in over 50\% of cases, exposing memorization-driven limitations; \emph{Math-RoB-Define} improves task interpretation with 1-shot prompting, while \emph{Math-RoB-Number} shows robustness to numerical and character substitutions.
Benchmark results show that, while larger closed-source models exhibit better instruction adherence, their improvements in robustness are marginal; in contrast, smaller open-source models are highly dependent on memorized training data, highlighting their vulnerability to overfitting.
In aggregate, we anticipate \emph{Math-RoB} will catalyze more research and refocus efforts on robust LLM reasoning.

% In our future work, 

\section{Limitations }

While our study provides some unique insights into reasoning robustness, there are still a few limitations to consider. The first limitation is about our hardware constraints and model scalability. Due to the hardware setup (dual RTX 4090 GPUs with 24GB VRAM each), our experiments focused on smaller models (< 8B parameters) with constrained search spaces (width < 6) and relatively shallow reasoning depths (30 steps). This means that we weren't able to explore larger models or deeper reasoning paths, so our findings may not fully extend to state-of-the-art models or broader search spaces. 
Also, our work may have model selection bias, i.e., the models we evaluated were selected based on their popularity on HuggingFace, which could introduce some bias. With fewer than five representative models in our study, we recognize that our conclusions may not capture the full diversity of mathematical ReFT across different architectures.

For the limitation on instruction-following dependency, our methodology relies on the models' ability to follow instructions, and we observed two areas for improvement. First is instruction sensitivity, where performance can decrease with less optimal prompts, particularly for tasks involving numerical substitution and operator replacement. Second is scalability, as larger models (over 7B parameters) tend to show better instruction compliance, although the reasons for this remain unclear. We also acknowledge that our framework doesn't yet provide a comprehensive analysis of instruction robustness across various architectures, which we plan to explore further in our future work.

\section{Ethical Considerations}
This study follows strict ethical guidelines, using only publicly available datasets without personal or sensitive information. We address potential biases in the models by analyzing training data and making efforts to enhance fairness and transparency. While focusing on improving LLM robustness, we also consider societal impacts, particularly preventing misuse and negative consequences. Our commitment is to balance technological advancement with social responsibility, contributing positively to the NLP field.

\bibliography{acl_latex}

\newpage
\onecolumn

\appendix
\section*{Appendix}

In this appendix, we present additional experimental results as well as detailed explanations of the evaluation process. Finally, we include a section on detailed related work relevant to this study.

\section{Evaluation Details}

\begin{figure}[ht]
    \centering
    \includegraphics[width=1\linewidth]{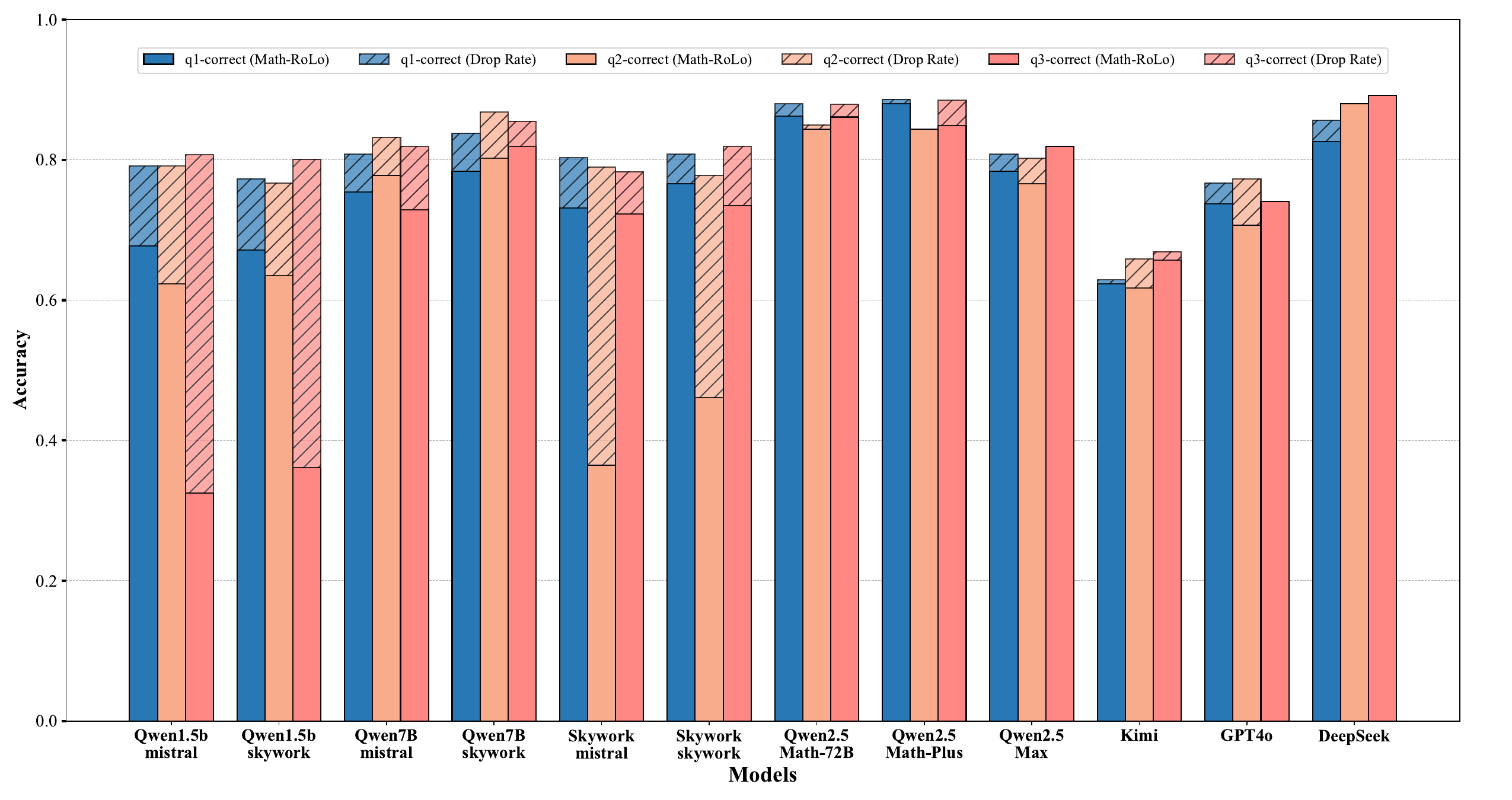}
    \caption{The response performance of different models on the Math-RoLo dataset for specific questions and their decline rate relative to Math500. The dashed line represents the decline rate.}
    \label{fig:enter-label}
\end{figure}

Figure \ref{fig:enter-label} presents the accuracy of various models on the Math-RoLo dataset, along with their drop rates compared to Math500. In Math500, different models achieve roughly the same accuracy for each question. Larger models (e.g., Qwen2.5 Math-Plus, GPT-4o, and DeepSeek) demonstrate higher accuracy, suggesting stronger generalization in mathematical reasoning. By contrast, smaller models (e.g., Qwen1.5B) show higher drop rates, indicating that the quality of the base model affects its ability to generalize. Increasing the scale—such as in Qwen7B—leads to lower drop rates, implying that larger models benefit from enhanced instruction-following capabilities, which in turn bolster reasoning robustness. Across different questions (q1, q2, and q3) within the same model, q2 and q3 exhibit higher drop rates, particularly for Qwen2.5- and Skywork-based LLMs, suggesting these models are more sensitive to input length and instruction following, and thus display weaker robustness.

\begin{table}[ht]
\renewcommand{\arraystretch}{1.5}
    \centering
    \begin{tabular}{|m{0.28\linewidth}|m{0.33\linewidth}|m{0.31\linewidth}|}
        \hline
        \textbf{Delete Question} & \textbf{Hallucinating problem-solving} & \textbf{Critical problem-solving} \\
        \hline
        
        {
        Determine the modulo \underline{\hspace{0.6em}} remainder of the following sum: 
        \[
        \begin{array}{c}
            1 + 2 + 3 + 4 + 5 + 6 + \\
            7 + 8 + 9 + 10 + 11 + 12.
        \end{array}
        \]
        }
        &
        {
        \includegraphics[width=0.1\linewidth]{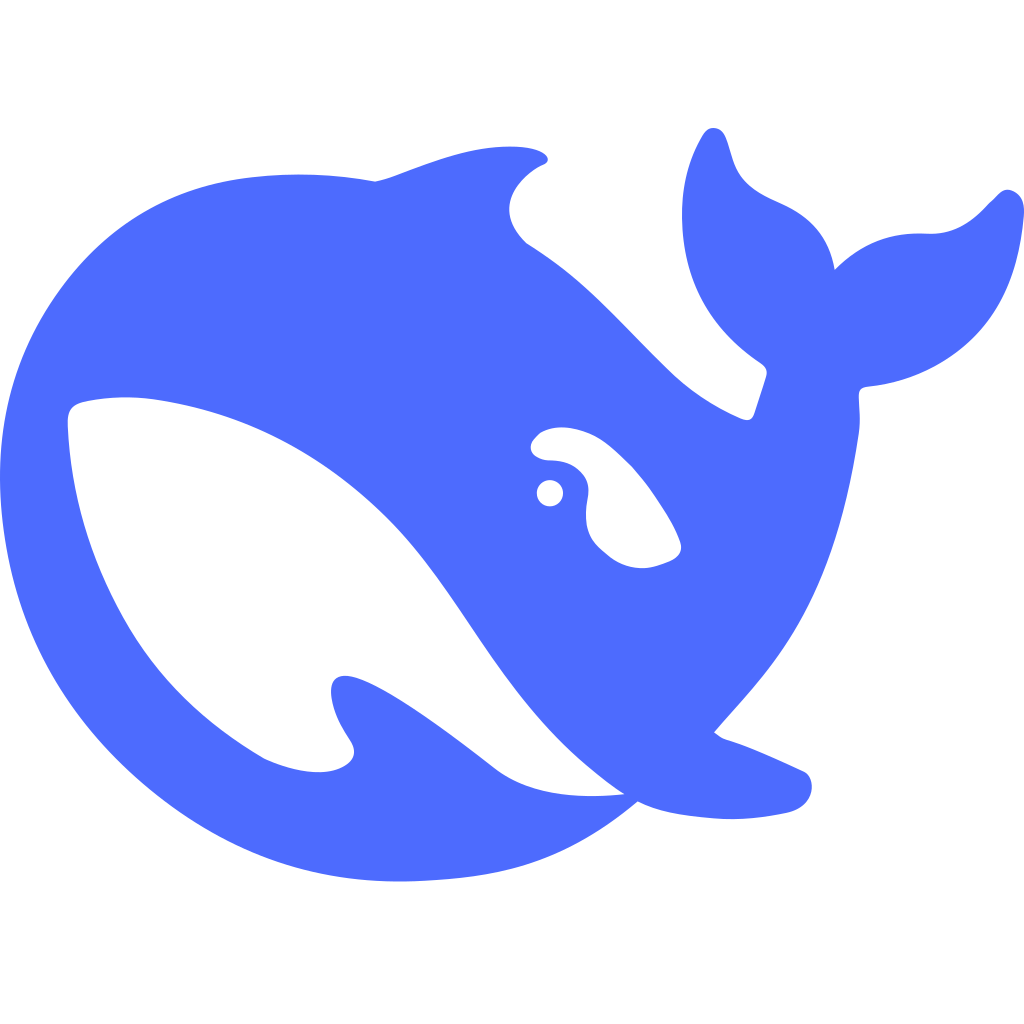} 
        To determine the modulo \textcolor{red}{7} remainder of the sum \(1 + 2 + 3 + 4 + 5 + 6 + 7 + 8 + 9 + 10 + 11 + 12\), follow these steps:
        
        1. **Calculate the sum of the numbers from 1 to 12:**
           The sum of the first \(n\) natural numbers is given by the formula:
           \[
           \text{Sum} = \frac{n(n + 1)}{2}
           \]
           For \(n = 12\):
           \[
           \text{Sum} = \frac{12 \times 13}{2} = 78
           \]
        
        2. **Find the remainder when 78 is divided by 7:**
           Divide 78 by 7:
           \[
           78 \div 7 = 11 \text{ with a remainder of } 1
           \]
           This is because \(7 \times 11 = 77\) and \(78 - 77 = 1\).
        
        3. **Final answer:**
           The modulo 7 remainder of the sum is \(\boxed{1}\).
        }
        & 
        {
        \includegraphics[width=0.1\linewidth]{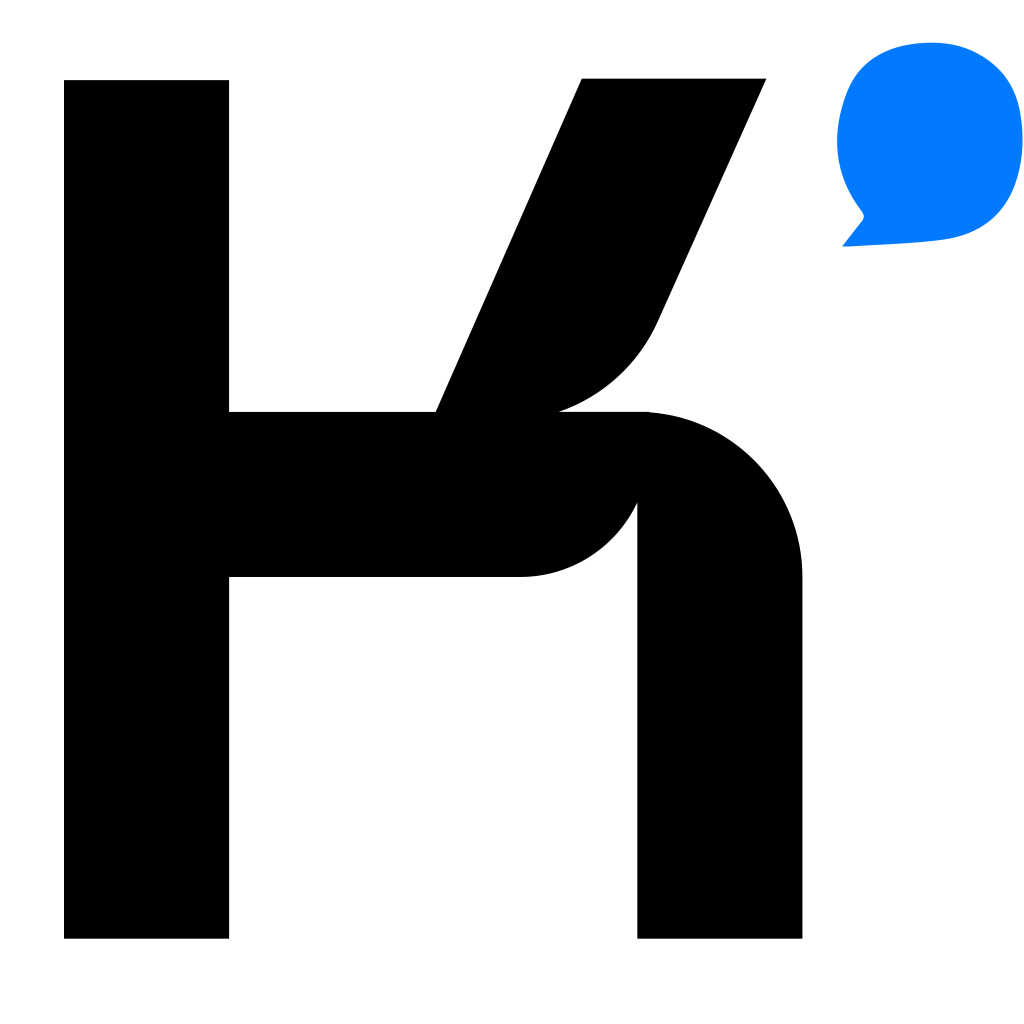} 
        To determine the modulo remainder of the sum \(1 + 2 + 3 + 4 + 5 + 6 + 7 + 8 + 9 + 10 + 11 + 12\), we will follow these steps:
        
        1. **Calculate the sum of the numbers:**
           The given sequence is an arithmetic series where the first term \(a = 1\) and the last term \(l = 12\). The number of terms \(n\) in the series can be found by:
           \vspace{-4mm} 
           \[
           n = l - a + 1 = 12 - 1 + 1 = 12
           \]
           
           \vspace{-4mm} 
           The sum \(S\) of the first \(n\) natural numbers is given by the formula:
           \vspace{-4mm} 
           \[
           S = \frac{n}{2} \times (a + l)
           \]

        \vspace{-2mm} 
           Substituting the values:
           \vspace{-4mm} 
           \[
           S = \frac{12}{2} \times (1 + 12) = 6 \times 13 = 78
           \]
           
        \vspace{-3mm} 
        2. **Determine the modulo remainder:**
           \textcolor{blue}{Since the problem does not specify a particular modulus, we will assume the most common modulus, which is 10 (for modulo 10).}
           To find the remainder when 78 is divided by \textcolor{blue}{10}, we perform the division:
            \vspace{-4mm} 
           \[
           78 \div 10 = 7 \text{ remainder } 8
           \]
           
            \vspace{-4mm} 
           Therefore, the remainder is 8.
        
        3. **Present the final answer in a box:** \(\boxed{8}\).
        }  % 修复：添加闭合的 }
        \\\hline
    \end{tabular}
    \caption{Instances of model hallucination and cases where the model identifies incomplete information and attempts to infer missing content. Here, we take the DeepSeek and Kimi models as examples for illustration.}
    \label{tab:modulo_remainder}
\end{table}

Table~\ref{tab:modulo_remainder} illustrates that, when faced with datasets containing missing information, most models rely on autoregressive mechanisms. These models generate complete responses by hallucinating missing data using maximum probability distributions. In contrast, a few models, such as Kimi and DeepSeek, explicitly identify the missing elements and hypothesize the absent information in order to construct a comprehensive response.

\section{Detailed Related Work}
\label{sect:detailed_related_work}

In this section, we provide a detailed literature review as an extension of the preliminaries (Sec.~\ref{sect:Related_Work}), including Benchmark for math reasoning (Appendix~\ref{sect:detailed_related_work_Benchmark}), research on CoT  (Appendix~\ref{sect:detailed_related_work_COT}), self-correction  (Appendix~\ref{sect:detailed_related_work_Self-Correction}) and reinforced Fine-Tuning (Appendix~\ref{sect:detailed_related_work_Reinforced}).

\subsection{Benchmarks for Math Reasoning}
\label{sect:detailed_related_work_Benchmark}
The GSM8K dataset \cite{cobbe2021training} is a widely used collection of 8.5K elementary math problems designed to test basic arithmetic and reasoning skills. The GSM-HARD dataset \cite{gao2023pal} builds on GSM8K by incorporating more complex and less common numbers. GSM-IC \cite{shi2023large} introduces irrelevant information to assess model performance under distractions, while GSM-PLUS \cite{li2024gsm} extends GSM8K with mathematical perturbations to test robustness against variations in problem formulation.

At the high-school level, the MATH dataset \cite{hendrycks2021measuring} provides 12.5K competition problems along with step-by-step solutions for a detailed evaluation. Math500 \cite{lightman2023let} is a curated subset of 500 problems from the MATH dataset designed for targeted evaluation across various mathematical topics. The Omni-MATH dataset \cite{gao2024omni} offers 4,428 competition-level problems across 33 subfields and 10 difficulty levels aimed at assessing advanced problem-solving skills.

For Chinese-language LLMs, the CMATH dataset \cite{wei2023cmath} includes 1.7K elementary-level math problems sourced from Chinese textbooks and exams. The SuperCLUE-Math6 dataset \cite{xu2024superclue} features 2,144 multi-step math problems designed to evaluate the reasoning abilities of Chinese LLMs.

The MathBench dataset \cite{liu2024mathbench} evaluates mathematical theory and problem-solving skills, covering topics from basic arithmetic to university-level mathematics. TheoremQA \cite{chen2023theoremqa} is the first dataset to focus on theorem-driven question-answering, containing 800 high-quality problems across mathematics, physics, engineering, and finance.

The Math2 dataset \cite{shevlizu1995math} combines two different mathematical skills from the MATH dataset into each problem, increasing diversity and challenge. MathAttack \cite{zhou2024mathattack} introduces semantic substitutions to test model robustness under adversarial conditions. The RobustMath dataset \cite{wang2023large} evaluates model performance under irrelevant contextual distractions, similar to GSM-IC but with a broader scope.

FrontierMath \cite{glazer2024frontiermath} introduces a benchmark with expert-crafted mathematical problems and automated verification to rigorously assess reasoning capabilities. MathPerturb \cite{jin2020bert} focuses on evaluating model robustness by introducing small perturbations to existing datasets, leading to different outcomes while maintaining high similarity to the original problems.

The MathWorld dataset \cite{shi2023large} is designed to test multi-step reasoning and the ability to handle complex, real-world mathematical problems. MathRobust \cite{li2023you} evaluates model performance under various perturbations, including semantic substitutions and irrelevant contextual distractions. Finally, MathEval \cite{zou2023meta} is a meta-benchmark that combines multiple existing datasets to provide a comprehensive evaluation of mathematical reasoning capabilities.

Together, these benchmarks cover a wide range of challenges in mathematical reasoning, from elementary arithmetic to advanced theoretical problems, domain specialization, robustness, and multi-step reasoning. They form the foundation for evaluating and improving the mathematical capabilities of LLMs, ensuring that models are equipped to handle a diverse set of problem types and complexities.

\subsection{Research on CoT}
\label{sect:detailed_related_work_COT}

The CoT prompting has emerged as a transformative paradigm for enhancing the reasoning capabilities of LLMs. By decomposing complex problems into intermediate reasoning steps, CoT mimics human cognitive processes, enabling LLMs to tackle tasks requiring multi-step logical inference, such as arithmetic, symbolic manipulation, and commonsense reasoning \cite{wei2022chain}. The foundational work by \citet{wei2022chain} demonstrated that explicit reasoning chains, embedded through few-shot examples, significantly improve model performance without fine-tuning. Subsequent research expanded this paradigm: \citet{kojima2022large} introduced zero-shot CoT, using generic prompts (e.g., "Let’s think step by step") to trigger step-by-step reasoning, while \citet{zhang2022automatic} automated CoT generation via question clustering (Auto-CoT), eliminating manual prompt engineering.

Recent advancements focus on improving CoT reliability and adaptability. \citet{wang2022self} proposed self-consistency decoding, aggregating multiple reasoning paths via majority voting to reduce errors. For domain-specific challenges, \citet{fei2023reasoning} applied CoT to implicit sentiment analysis through a structured three-step framework (THOR), achieving 50\% gains in zero-shot settings. Hybrid approaches integrate CoT with search algorithms: \citet{zhao2024marco} combined Monte Carlo Tree Search (MCTS) with reflection mechanisms (MARCO) to iteratively refine reasoning paths, and \citet{xie2024monte} leveraged MCTS to explore and rank optimal solutions. Reinforcement learning further enhanced CoT through methods like Reinforced Fine-Tuning (ReFT) \cite{trung2024reft}, which optimizes stepwise reasoning via reward-based path evaluation.

Despite progress, CoT faces two critical limitations. First, its efficacy diminishes in non-mathematical domains like natural language understanding, where contextual ambiguity persists \cite{sprague2024cot}. Second, computational overhead from generating lengthy reasoning chains hinders real-time applications. To address efficiency \cite{jing2021meta}, \citet{yao2024tree} introduced Tree-of-Thought structures that prune redundant paths while maintaining performance. Future directions include meta-reasoning frameworks like Meta-CoT \cite{zou2023meta}, which explicitly models reasoning processes, and cross-domain adaptations such as hierarchical decomposition strategies \cite{zhao2023improving}.

\subsection{Self-Correction}
\label{sect:detailed_related_work_Self-Correction}
Self-correction in LLMs has emerged as a promising direction for improving output quality through iterative refinement. This capability enables models to identify and rectify errors during reasoning or generation, enhancing their reliability for real-world applications.

Recent work explores various self-correction paradigms. \citet{shinn2023reflexion} introduce \emph{Reflexion}, where language agents employ verbal reinforcement learning to refine outputs through reflective feedback. Similarly, \citet{madaan2024self} propose Self-Refine, which iteratively generates initial outputs, provides multi-dimensional self-feedback, and revises responses until convergence. Building on this, \citet{ye2023selfee} develop SelFee by fine-tuning LLA MA with ChatGPT-generated training instances, demonstrating improved instruction-following through self-feedback mechanisms.

For complex reasoning tasks, \citet{kim2023language} present Recursively Criticizes and Improves (RCI), a prompting framework that enables LLMs to recursively critique and improve their solutions. Addressing error localization challenges, \citet{tyen2023llms} propose Backtracking (BT), which guides correction by identifying initial error positions in problem-solving sequences.

When intrinsic correction proves insufficient, external feedback integration shows promise. \citet{paul2023refiner} employs a separate critic model to generate structured feedback for iterative refinement, while \citet{akyurek2023rl4f} develops a collaborative framework where specialized critique generators optimize reasoning model performance through reinforcement learning.

Notably, \citet{huang2022large} demonstrate LLMs' capacity for gradual self-improvement through output comparison, though subsequent work by \citet{huang2023large} highlights inherent limitations in unaided self-correction for complex reasoning tasks.

\subsection{Reinforced Fine-Tuning}
\label{sect:detailed_related_work_Reinforced}
Reinforcement learning from human feedback (RLHF) is critical for various areas, such as computer vision \cite{jing2021turning,pinto2023tuning,jing2023deep} and natural language processing \cite{lightman2023let}.
In particular, recent advances have enhanced language model alignment by integrating supervised fine-tuning with preference-based optimization. This approach addresses the limitations of purely supervised methods by leveraging reward models trained on human preferences \citep{ouyang2022training}.

The RLHF paradigm builds on pairwise preference learning \citep{christiano2017deep}, first adapted for LLMs by \citet{ziegler2019fine}. Subsequent work demonstrated RLHF's effectiveness in text generation tasks through proximal policy optimization (PPO) \citep{stiennon2020learning,schulman2017proximal}. Theoretical analyses further revealed RLHF's implicit reward modeling capabilities \citep{perez2022discovering}.

Advances in reward modeling have improved the RLHF stability, with scaling laws highlighting the importance of preference data quality \citep{gao2023scaling}. Hybrid human-AI annotation approaches \citep{touvron2023llama} and adversarial training methods \citep{ramamurthy2022reinforcement} have further enhanced robustness. Standardized evaluation protocols now enable systematic comparisons \citep{borgeaud2022improving}.

Researchers have developed alternatives to traditional PPO-based optimization. \citet{bai2022training} introduced self-supervised reinforcement learning with AI-generated feedback, while \citet{rafailov2023direct} proposed Direct Preference Optimization (DPO), eliminating explicit reward modeling through gradient-based policy learning. Hybrid frameworks integrating adversarial learning with RL fine-tuning \citep{ramamurthy2022reinforcement} demonstrate improved sample efficiency.

Key challenges persist in reward function design \citep{ziegler2019fine} and policy optimization stability \citep{schulman2017proximal}. Addressing these requires advances in theoretical understanding of high-dimensional policy spaces \citep{mnih2016asynchronous} and improved human-AI collaboration frameworks \citep{ouyang2022training}.

\end{document}